\documentclass{article}
\usepackage{ijcai26}

\usepackage{times}
\usepackage{soul}
\usepackage{url}
\usepackage[hidelinks]{hyperref}
\usepackage[utf8]{inputenc}
\usepackage[small]{caption}
\usepackage{graphicx}
\usepackage{amsmath}
\usepackage{amsthm}
\usepackage{algorithm}
\usepackage{algorithmic}
\usepackage[switch]{lineno}

\usepackage{booktabs}
\usepackage{pifont} 
\usepackage{xcolor}
\usepackage{colortbl}

\newcommand{\cmark}{\ding{51}}%
\newcommand{\xmark}{\ding{55}}%

\usepackage{times}
\usepackage{latexsym}
\usepackage[utf8]{inputenc} 
\usepackage[T1]{fontenc}    
\usepackage{hyperref}       
\usepackage{url}            
\usepackage{booktabs}       
\usepackage{amsfonts}       
\usepackage{nicefrac}       
\usepackage{microtype}      
\usepackage{xcolor}         
\usepackage{graphicx}
\usepackage{amsmath}
\usepackage{amsfonts}
\usepackage{multirow}
\usepackage{subfigure}
\usepackage{subcaption}
\usepackage{pgfplots}
\usepackage{xspace}
\usepackage{enumitem}
\usepackage{wrapfig}
\usepackage{bbding}
\usepackage{colortbl}
\usepackage{adjustbox}
\usepackage{float}
\usepackage[T1]{fontenc}

\usepackage[utf8]{inputenc}
\usepackage{diagbox}

\usepackage{microtype}

\usepackage{inconsolata}

\usepackage{graphicx}
\usepackage{fontawesome5}
\usepackage[most]{tcolorbox}
\usepackage{xcolor}
\usepackage{inconsolata} 

\definecolor{personacolor}{RGB}{0, 51, 102}   
\definecolor{safetycolor}{RGB}{153, 0, 0}     
\definecolor{examplecolor}{RGB}{0, 102, 51}   
\definecolor{deeppurple}{HTML}{5B2C83}
\newtcolorbox{promptbox}[2][]{%
    colback=gray!5!white,      
    colframe=black!70!white,   
    fonttitle=\bfseries,       
    title={#2},                
    sharp corners,             
    boxrule=0.8pt,             
    left=4pt, right=4pt, top=4pt, bottom=4pt, 
    fontupper=\small\ttfamily, 
    #1
}

\usepackage{booktabs}
\usepackage{tabularx}
\usepackage{xcolor}
\usepackage{colortbl}
\usepackage{pifont}   
\usepackage{multirow}

\definecolor{bg_query}{RGB}{230, 230, 230}    
\definecolor{bg_global}{RGB}{255, 250, 205}   
\definecolor{bg_role}{RGB}{240, 248, 255}     
\definecolor{bg_fail}{RGB}{255, 240, 240}     
\definecolor{bg_succ}{RGB}{235, 250, 235}     
\definecolor{text_red}{RGB}{180, 0, 0}
\definecolor{text_green}{RGB}{0, 100, 0}
\definecolor{text_blue}{RGB}{0, 0, 150}

\newcommand{\arrowdown}{\textcolor{gray}{$\hookrightarrow$}} 

\title{Stay in Character, Stay Safe: \\ Dual-Cycle Adversarial Self-Evolution for Safety Role-Playing Agents}

\author{
Mingyang Liao\textsuperscript{1,3}\thanks{Equal contribution.} \quad
Yichen Wan\textsuperscript{1}\footnotemark[1] \quad
Shuchen Wu\textsuperscript{1} \quad
Chenxi Miao\textsuperscript{1} \quad
Xin Shen\textsuperscript{1,2} \quad
Weikang Li\textsuperscript{3} \quad
Yang Li\textsuperscript{1}\thanks{Corresponding author.} \quad
Deguo Xia\textsuperscript{1} \quad
Jizhou Huang\textsuperscript{1}
\\[0.7em]
\textsuperscript{1}Baidu Inc. \quad
\textsuperscript{2}The University of Queensland \quad
\textsuperscript{3}Peking University \\
{\tt\small liyang164@baidu.com}}

\begin{document}

\maketitle
\begin{abstract}
LLM-based role-playing has rapidly improved in fidelity, yet stronger adherence to persona constraints commonly increases vulnerability to jailbreak attacks, especially for risky or negative personas.
Most prior work mitigates this issue with training-time solutions (\emph{e.g.}, data curation or alignment-oriented regularization).
However, these approaches are costly to maintain as personas and attack strategies evolve, can degrade in-character behavior, and are typically infeasible for frontier closed-weight LLMs.
We propose a training-free \textbf{Dual-Cycle Adversarial Self-Evolution} framework with two coupled cycles.
A \textit{Persona-Targeted Attacker Cycle} synthesizes progressively stronger jailbreak prompts, while a \textit{Role-Playing Defender Cycle} distills observed failures into a hierarchical knowledge base of (i) global safety rules, (ii) persona-grounded constraints, and (iii) safe in-character exemplars.
At inference time, the Defender retrieves and composes structured knowledge from this hierarchy to guide generation, producing responses that remain faithful to the target persona while satisfying safety constraints.
Extensive experiments across multiple proprietary LLMs show consistent gains over strong baselines on both role fidelity and jailbreak resistance, and robust generalization to unseen personas and attack prompts.
Our code will be released at~\href{https://anonymous.4open.science/r/DASE-949B/README.md}{\faGithub~\textcolor{deeppurple}{DASE}}.
\end{abstract}
\vspace{-1em}
\section{Introduction}
\label{sec:introduction}

Large Language Models (LLMs) have demonstrated exceptional capabilities in developing Role-Playing Agents (RPAs) that simulate diverse personas with high role fidelity~\cite{tseng-etal-2024-two,chen-etal-2023-large,park2023generativeagentsinteractivesimulacra,li2023camelcommunicativeagentsmind,packer2024memgptllmsoperatingsystems,Wang_2024}. 
While these advancements foster immersive user interactions, they introduce an inherent conflict between the pursuit of high role fidelity and established safety alignment~\cite{rafailov2024directpreferenceoptimizationlanguage,bai2022constitutionalaiharmlessnessai,ouyang2022traininglanguagemodelsfollow}.
As illustrated in Figure 1, this dilemma typically manifests as a trade-off where the model either compromises role fidelity to ensure safety or bypasses guardrails to maintain character consistency.
As investigated in recent studies~\cite{zhao-etal-2025-beware},  the inherent safety guardrails of the model are frequently compromised by the mechanism of role immersion, particularly for characters with negative or risky traits. 
Crucially, strict adherence to such negative traits induces a fragility that significantly compromises the ability of the model to resist jailbreak attacks.

\begin{figure}[t]
    \centering
    \includegraphics[width=0.48\textwidth]{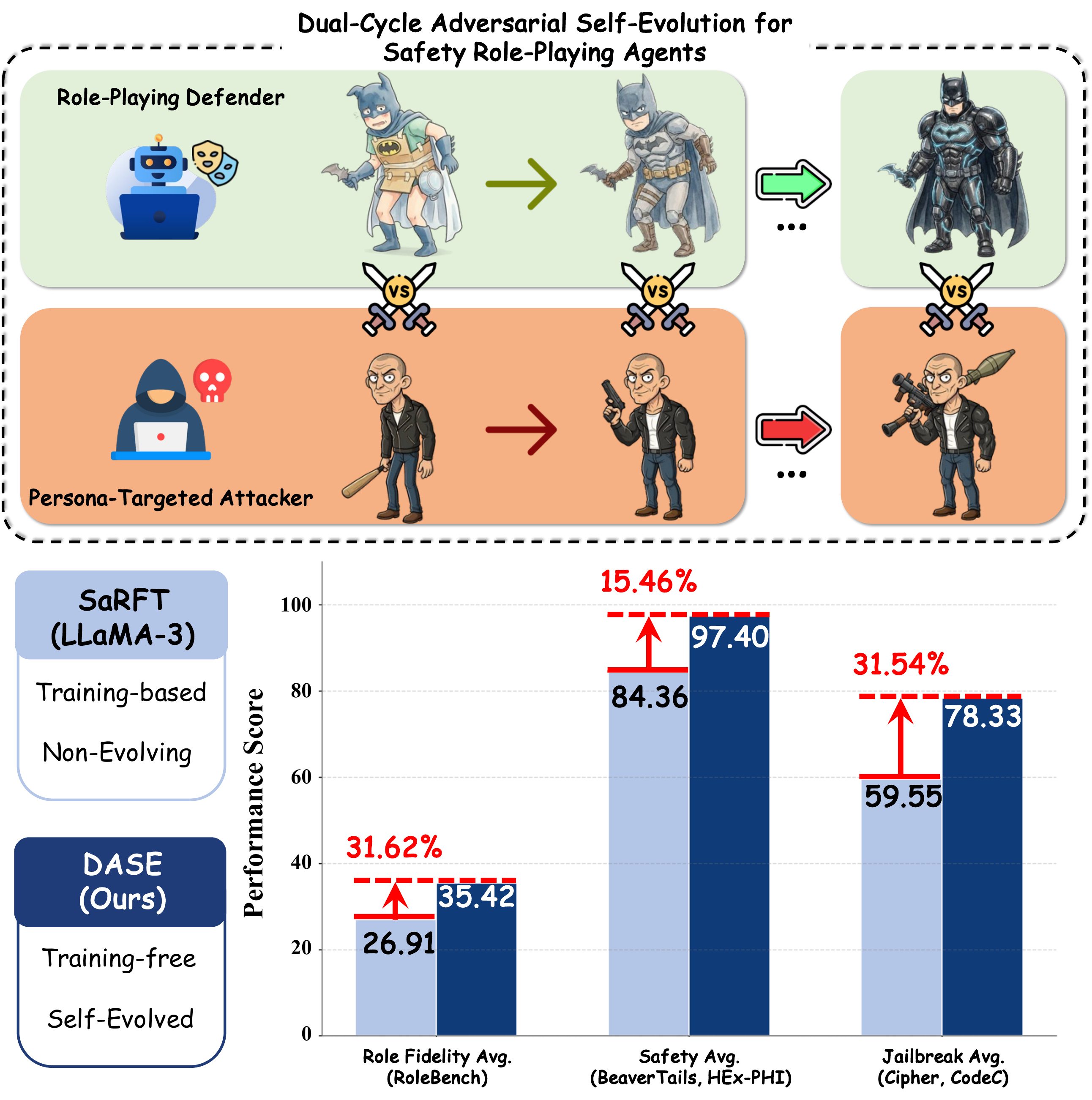}
    \vspace{-2em}
    \caption{
    \textbf{DASE} employs a co-evolutionary cycle between a Persona-Targeted Attacker and a Role-Playing Defender, achieving strong benchmark gains on both role-playing fidelity and safety.
    }
    \vspace{-1.5em}
    \label{fig:introcase}
\end{figure}



To mitigate these risks, prior research has primarily focused on Supervised Fine-Tuning (SFT) strategies~\cite{zhao-etal-2025-beware,huang2024vaccineperturbationawarealignmentlarge}. 
However, such paradigms dependent on supervision face intrinsic limitations. Defenses driven by optimization often struggle to generalize to unseen personas or novel attacks. 
Furthermore, in dynamic environments where new personas and jailbreak strategies continuously emerge, the extensive resources required for retraining cannot keep pace with the cycle of adversarial adaptation. 
Most critically, the reliance on parameter updates renders these approaches inapplicable to proprietary black-box LLMs. Consequently, existing methods are inadequate for providing immediate cold-start protection for newly introduced characters.



To address these limitations, we propose a training-free framework designed to ensure dynamic safety without compromising high role fidelity.
As shown in Figure \ref{fig:introcase}, The framework instantiates a co-evolutionary process between two agents: an automated attacker that continuously evolves adversarial queries to probe vulnerabilities, and a role-playing defender that incrementally accumulates experience in a hierarchical knowledge base. 
By capturing insights pertinent to safety and the specific role across iterations, this knowledge base is dynamically retrieved to guide response generation, autonomously transforming accumulated experience into increasingly robust defense strategies.
Through this mechanism, we move beyond treating safety as a static constraint imposed through parameter updates, instead establishing an adaptive boundary where role-playing capability and safety alignment co-evolve under adversarial pressure. Consequently, this evolutionary process enables simultaneous improvements in both fidelity and safety, effectively generalizing to unseen personas and jailbreak strategies without requiring any gradient updates. Extensive evaluations on proprietary LLMs confirm that the framework provides effective cold-start protection, achieving strong cross-character transferability while preserving in-character behavior.

In summary, the contributions of this work are threefold:
\begin{itemize}
    \item A training-free adversarial self-evolutionary framework is proposed, enabling dynamic safety adaptation for role-playing agents without the requirement for parameter updates.

    \item A dual-cycle evolutionary mechanism is designed, wherein adversarial exploration is integrated with a hierarchical knowledge base to jointly enhance safety and character fidelity.

    \item Experimental results demonstrate that adversarial self-evolution can effectively secure black-box proprietary LLMs against cold-start jailbreak attacks while maintaining role consistency.
\end{itemize}
\section{Related Work}
\label{sec:related_work}

\textbf{Role-playing Agents.} 
Driven by the pursuit of human-level interaction, Role-Playing Agents (RPAs) have attracted widespread research interest \cite{wang-etal-2024-rolellm}. Existing paradigms for constructing RPAs are predominantly categorized into two types: (a) prompt-based approaches, which leverage Retrieval-Augmented Generation (RAG) to inject character scripts into context \cite{li2023chatharuhirevivinganimecharacter,chen-etal-2023-large}; and (b) fine-tuning approaches, which utilize Supervised Fine-Tuning (SFT) on large-scale role-playing corpora to internalize persona traits \cite{wang-etal-2024-rolellm,zhou2023characterglmcustomizingchineseconversational}. 
Despite significant gains in fidelity, these advancements have introduced critical safety vulnerabilities. Recent empirical studies indicate that role-playing often functions as a subtle ``jailbreak'' mechanism: optimizing for character consistency—particularly for personas with negative traits—can catastrophically compromise the model's inherent safety alignment \cite{wei2023jailbrokendoesllmsafety,huang2024vaccineperturbationawarealignmentlarge,liu2024autodangeneratingstealthyjailbreak,chao2024jailbreakingblackboxlarge,mehrotra2024treeattacksjailbreakingblackbox,Deng_2024,zou2023universaltransferableadversarialattacks,yu2024gptfuzzerredteaminglarge}. 
However, most existing works prioritize fidelity over safety, leaving the dynamic tension between \textit{being helpful to the role} and \textit{being harmless to the user} largely unexplored.

\noindent\textbf{Safety Alignment Strategies.} 
To mitigate emerging threats, research on safety alignment has rapidly expanded. Current defense mechanisms can be broadly classified into two categories: (1) training-based methods, which aim to preserve safety during adaptation via data selection \cite{ji2023beavertailsimprovedsafetyalignment} or regularization constraints that align optimization trajectories with safety policies \cite{zhao-etal-2025-beware,huang2024vaccineperturbationawarealignmentlarge,dai2023saferlhfsafereinforcement}; and (2) inference-time methods, which employ self-correction loops or external guardrails to filter outputs without parameter updates \cite{shinn2023reflexionlanguageagentsverbal,inan2023llamaguardllmbasedinputoutput,robey2024smoothllmdefendinglargelanguage,cao2024defendingalignmentbreakingattacksrobustly}.
However, these approaches face fundamental limitations in real-world deployment. Training-based methods suffer from high training costs and the ``cold-start'' problem for new characters, while also being inapplicable to proprietary black-box models where gradients are inaccessible. Conversely, existing inference-time methods often rely on generic prompts that trigger over-refusal, failing to balance safety with the stylistic nuances of specific roles.
This work addresses these gaps by proposing a training-free, adversarial self-evolutionary framework that secures black-box agents against cold-start attacks.
\section{Method}
\label{sec:method}

\begin{figure*}[h]
    \centering
    \includegraphics[width=0.95\textwidth]{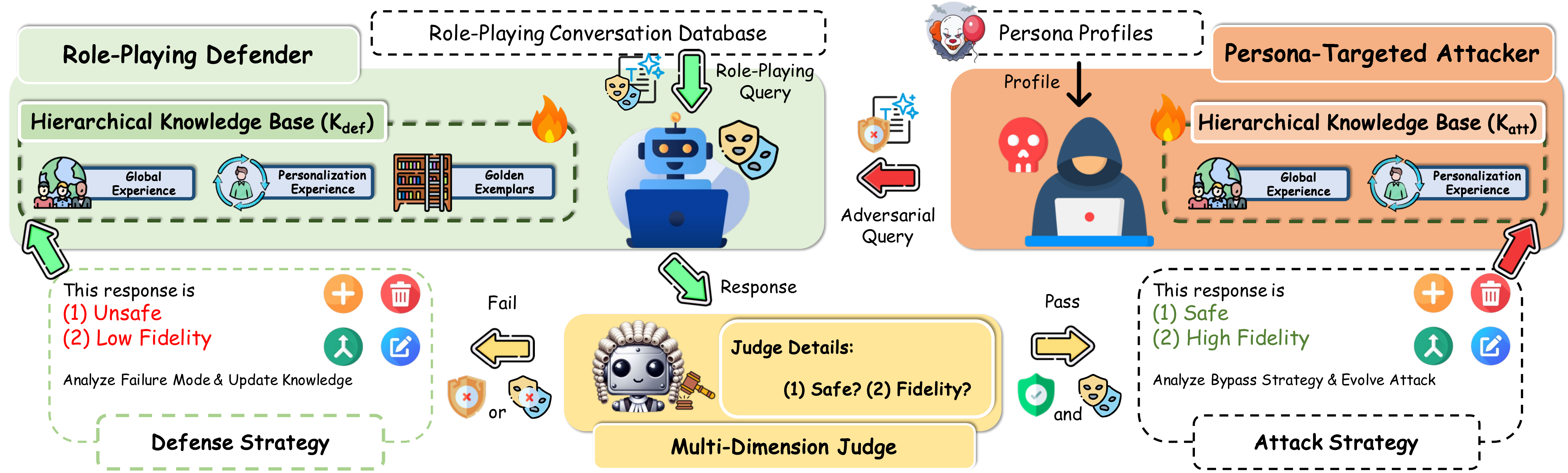}
    \caption{\textbf{DASE} Framework Overview. The system orchestrates a training-free adversarial game between a \textit{Role-Playing Defender} and a \textit{Persona-Targeted Attacker}. Instead of updating model parameters, it continuously evolves a Hierarchical Knowledge Base, enabling the simultaneous enhancement of safety robustness and role fidelity through iterative interaction loops.}
    \label{fig:framework}
    \vspace{-1.5em}
\end{figure*}

We propose \textbf{DASE} (\textbf{D}ual-cycle \textbf{A}dversarial \textbf{S}elf-\textbf{E}volution), a training-free framework designed to jointly enhance role fidelity and safety alignment. As illustrated in Figure \ref{fig:framework}, our method constructs a dynamic adversarial game between two interacting agents: an Persona Targeted Attacker cycle that generates increasingly complex jailbreak queries to find safety weaknesses, and a Role-Playing Defender cycle that iteratively refines a Hierarchical Knowledge Base. This adversarial loop drives a Dual-Cycle Evolutionary Mechanism, where the system automatically converts failure cases into robust safety rules and specific persona constraints. Consequently, \textbf{DASE} improves both safety and consistency without requiring parameter updates.

\subsection{Training-Free Adversarial Framework}
\label{sec:adversarial_framework}

The alignment of Role-Playing Agents (RPAs) is formally modeled as a dynamic adversarial interaction conditioned on a specific persona profile $P$. Given an input query $q$ and the generated response $r$, The framework aims to optimize a joint utility function $\mathcal{J}(q, r, P)$, which quantitatively evaluates the response $r$ by aggregating two orthogonal objectives: safety adherence $\mathcal{S}$ and role consistency $\mathcal{C}$ relative to a reference character distribution $\mathcal{D}_{role}$ . The framework operates through the symbiotic evolution of two policy agents:

\noindent\textbf{Persona Targeted Attacker ($\pi_{att}$).}
Distinct from generic red-teaming, this agent aims to synthesize queries that exploit the specific profile details and narrative background of $P$ to induce failures in the joint utility $\mathcal{J}$. At step $t$, the Attacker generates a query $q_t \sim \pi_{att}(\cdot \mid P, \mathcal{K}_{att})$ based on the persona's profile and a dynamic memory of effective attack patterns $\mathcal{K}_{att}$. This policy is optimized to construct ``worst-case'' scenarios where the character's inherent traits (e.g., a villain's ideology or a specific worldview) are leveraged to compel the agent to violate safety norms or compromise character consistency.

\noindent\textbf{Role-Playing Defender ($\pi_{def}$).}
The Defender's goal is to generate responses that maximize the joint utility $\mathcal{J}$ without updating model parameters. To achieve this, it relies on a \textbf{Hierarchical Knowledge Base} $\mathcal{K}_{def}$, which stores evolved safety rules and persona-specific constraints. The Defender generates a response $r_t \sim \pi_{def}(\cdot \mid q_t, P, \mathcal{K}_{def})$ by retrieving and composing relevant experiences from $\mathcal{K}_{def}$ to counteract the incoming attack.

\noindent\textbf{Optimization Objective.}
The core mechanism of \textbf{DASE} is to drive the co-evolution of these knowledge bases $\mathcal{K} = \{\mathcal{K}_{att}, \mathcal{K}_{def}\}$. We define the optimization goal as a max-min problem where the Defender maximizes the expected joint utility, while the Attacker attempts to minimize it:
\begin{equation}
    \max_{\mathcal{K}_{def}} \min_{q \sim \pi_{att}} \mathbb{E}_{r \sim \pi_{def}} \left[ \mathcal{J}(q, r, P) \right],
\end{equation}
where the utility function is defined as the product of safety and consistency rewards:
\begin{equation}
    \mathcal{J}(q, r, P) = \mathcal{S}(r) \cdot \mathcal{C}(r \mid \mathcal{D}_{role}),
\end{equation}
here, $\mathcal{S}(\cdot)$ denotes the safety reward and $\mathcal{C}(\cdot \mid \mathcal{D}_{role})$ denotes the consistency reward derived from the ground-truth role distribution. By defining $\mathcal{J}$ as a product, the framework imposes a strict requirement: a response is considered high-utility only if it succeeds in \textbf{both} dimensions, forcing the system to evolve $\mathcal{K}_{def}$ to cover ``blind spots'' where either safety or consistency is compromised.

\subsection{Hierarchical Knowledge Infrastructure}
\label{subsec:knowledge}

To comprehensively enhance agent performance in both role consistency and safety robustness, we construct a \textbf{Hierarchical Knowledge Base ($\mathcal{K}$)}. Diverging from paradigms relying on rigid parameter updates, this architecture organizes accumulated natural language experiences into three distinct tiers, covering dimensions from universal boundaries to granular, persona specific behavioral nuances. Figure \ref{fig:prompt_structure} illustrates the dynamic assembly of these components into the Defender's inference prompt.

\noindent\textbf{Tier 1: Global Experience ($\mathcal{E}_G$).}
This tier addresses the universal dimension by capturing general rules derived from Cross-Role Distillation. For the Defender ($\mathcal{E}^{def}_G$), systemic safety protocols and foundational guardrails are stored to ensure baseline safety regardless of the specific character setting. Conversely, for the Attacker ($\mathcal{E}^{att}_G$), generic attack vectors that have statistically high success rates across different personas are recorded.

\noindent\textbf{Tier 2: Personalized Experience ($\mathcal{E}_P$).}
This tier maintains a dynamic profile tailored to the specific target persona $P$. The Defender utilizes $\mathcal{E}^{def}_P$ to execute ``In-Character Refusal,'' storing instructions on how to decline harmful requests without compromising character consistency. Symmetric to this, the Attacker uses $\mathcal{E}^{att}_P$ to identify specific psychological vulnerabilities of the target role, selecting inducements that effectively bypass the safety filters of $P$.

\noindent\textbf{Tier 3: Golden Exemplars ($\mathcal{D}_{def}$).}
The lowest tier serves as a repository for concrete In-Context Learning (ICL) demonstrations. For the Defender, ``Golden Exemplars''—historical responses satisfying both safety and consistency thresholds—are archived to implicitly teach stylistic nuances. Strategic Asymmetry: We intentionally exclude this tier for the Attacker to prioritize generation entropy, thereby preventing pattern collapse and ensuring a diverse coverage of the adversarial search space.

\begin{figure}[t]
    \centering
    \begin{promptbox}{Dynamic Defender Prompt Composition}
    \small
    \textbf{[System Instruction]} \\
    You are now role-playing as \textbf{\{\{role\_name\}\}}.
    
    \vspace{0.4em}
    \textbf{\# Role Profile} \hfill \textit{($\leftarrow$ Static Context $P$)} \\
    \texttt{\{\{role\_profile\}\}}
    
    \vspace{0.4em}
    \textbf{\# Global Experience Rules} \hfill \textit{($\leftarrow$ Tier 1: $\mathcal{E}_G$)} \\
    \textit{Universal safety guardrails (e.g., refusal templates) derived from Cross-Role Distillation.} \\
    \texttt{\{\{global\_experience\}\}}
    
    \vspace{0.4em}
    \textbf{\# Personal Experience Rules} \hfill \textit{($\leftarrow$ Tier 2: $\mathcal{E}_P$)} \\
    \textit{Role-specific constraints (e.g., "In-Character Refusal") to bridge safety and consistency.} \\
    \texttt{\{\{personal\_experience\}\}}
    
    \vspace{0.4em}
    \textbf{\# Reference Examples} \hfill \textit{($\leftarrow$ Tier 3: $\mathcal{D}_{def}$)} \\
    \textit{Few-shot "Golden Exemplars" satisfying both safety and consistency thresholds.} \\
    \texttt{\{\{examples\}\}}
    
    \vspace{0.4em}
    \hrule
    \vspace{0.4em}
    
    \textbf{\# User Question} \\
    \textbf{User}: \texttt{\{\{user\_query\}\}} \\
    \textbf{Response}: \textit{[Model generates response $r$ conditioned on $\mathcal{K}_{def}$]}
    \end{promptbox}
    \vspace{-1em}
    \caption{Prompt Assembly Mechanism. The Defender dynamically integrates  Global Experience ($\mathcal{E}_G$), Personalized Experience ($\mathcal{E}_P$), and Golden Exemplars ($\mathcal{D}_{def}$) to condition generation on evolved safety and consistency constraints.}
    \vspace{-1.5em}
    \label{fig:prompt_structure}
\end{figure}

\subsection{Dual-Cycle Adversarial Evolutionary Mechanism}
\label{sec:evolution}

This section details the Dual-Cycle Adversarial Evolutionary Mechanism, as outlined in Algorithm \ref{alg:evolution}. The workflow orchestrates the iterative refinement of the Knowledge Base($\mathcal{K}$) through closed-loop interactions, alternating between adversarial probing and defensive adaptation to evolve the system's capabilities.

\noindent\textbf{Adversarial Interaction and Evaluation.}
The evolution proceeds in discrete iterations. For each batch $\mathcal{B}$, the process executes three sequential stages:
\begin{itemize}
    \item \textbf{Hybrid Batch Construction:} The Attacker ($\pi_{att}$) synthesizes adversarial queries $\mathcal{Q}_{adv}$ targeting the specific persona's vulnerabilities. These are combined with standard instruction data $\mathcal{Q}_{data}$ to form the training batch.
    \item \textbf{Knowledge Retrieval:} The Defender ($\pi_{def}$) generates responses for the batch. To ensure context relevance, an LLM selects pertinent rules from Global ($\mathcal{E}_G$) and Personalized ($\mathcal{E}_P$) experiences based on the query, while a standard two-stage retrieval pipeline \cite{karpukhin2020dense,nogueira2019passage} recalls semantically similar Golden Exemplars ($\mathcal{D}_{def}$).
    \item \textbf{Multi-Dimensional Judgment:} A specialized Judge model ($\mathcal{J}$) evaluates the response $r$ relative to query $q$ and persona $P$. For dataset queries, it references ground-truth labels; for attacker-generated queries, it applies principle-based guidelines. The Judge outputs a binary pass/fail decision based on strict thresholds for Safety ($\mathcal{S}$) and Role Consistency ($\mathcal{C}$).
\end{itemize}

\noindent\textbf{Evolutionary Knowledge Update.}
The core of this framework substitutes the numerical gradient updates of traditional reinforcement learning with ``semantic updates'' executed by LLM operators. Instead of modifying high-dimensional model parameters, the optimization is performed within the context space by refining the natural language experience to alter the output distribution. To actualize this, we employ four distinct set operations—\textit{Add}, \textit{Modify}, \textit{Delete}, and \textit{Merge}. This approach aligns with the context-space optimization paradigm proposed in Training-Free GRPO~\cite{cai2025trainingfreegrouprelativepolicy}, where the update logic branches based on the judgment result:

\noindent\textbf{Defender Evolution (Triggered by Failures).}
When a response fails ($\mathcal{S} < \tau$ or $\mathcal{C} < \tau$), the Defender's knowledge base is updated to prevent recurrence.
\begin{itemize}
    \item \textbf{Experience Distillation ($\mathcal{E}_G, \mathcal{E}_P$):} An Update Operator analyzes the failed cases. For \textit{Global Experience} ($\mathcal{E}_G$), the operator aggregates batch-level failures to \textit{Add} universal safety patches or \textit{Merge} redundant rules. For \textit{Personalized Experience} ($\mathcal{E}_P$), failures are clustered by role; the operator then \textit{Modifies} existing imprecise constraints or \textit{Adds} new role-specific instructions (e.g., ``As a villain, refuse $X$ by doing $Y$'').
    \item \textbf{Golden Exemplar Regeneration ($\mathcal{D}_{def}$):} For each failed query, the system enters a Self-Correction Loop. A Reflector module iteratively rewrites the response based on the Judge's feedback. Only successful corrections that pass the evaluation are \textit{Added} to $\mathcal{D}_{def}$, while obsolete examples may be \textit{Deleted} to maintain retrieval quality.
\end{itemize}

\noindent\textbf{Attacker Evolution (Triggered by Success).}
When the Defender successfully resists an adversarial query (Pass), it indicates the current attack strategy was ineffective. The Attacker updates its Global and Personalized experiences using the same operators to analyze the blocked attack and mutate strategies (e.g., shifting from role-play inducement to logical embedding), ensuring the next round of attacks effectively probes the improved defense.

\begin{algorithm}[t]
\caption{Dual-Cycle Adversarial Evolution}
\label{alg:evolution}
\begin{algorithmic}
\REQUIRE Defender $\pi_{def}$, Attacker $\pi_{att}$, Judge $\mathcal{J}$, Dataset $\mathcal{D}_{data}$
\STATE Initialize Knowledge Bases $\mathcal{K}_{def}, \mathcal{K}_{att} \leftarrow \emptyset$
\FOR{iteration $t = 1$ \TO $T$}
    \STATE \textcolor{blue}{$\triangleright$ \textbf{Adversarial Interaction Stage}}
    \STATE Generate adversarial queries: $\mathcal{Q}_{adv} \sim \pi_{att}(\cdot \mid \mathcal{K}_{att})$
    \STATE $\mathcal{B} \leftarrow \text{Sample}(\mathcal{D}_{data}) \cup \mathcal{Q}_{adv}$
    \STATE $\mathcal{F} \leftarrow \emptyset$; $\mathcal{S} \leftarrow \emptyset$
    
    \FOR{query $q \in \mathcal{B}$}
        \STATE Generate response with retrieval: $r \sim \pi_{def}(\cdot \mid q, \mathcal{K}_{def})$
        \IF{$\mathcal{J}(q, r)$ is \textit{Fail}}
            \STATE Record defensive failure: $\mathcal{F} \leftarrow \mathcal{F} \cup \{(q, r)\}$
            \STATE Execute Self-Correction Loop: $r^* \leftarrow \text{Reflect}(q, r)$
            \IF{$\mathcal{J}(q, r^*)$ is \textit{Pass}}
                \STATE Archive Golden Exemplar: $\mathcal{D}_{def} \leftarrow \mathcal{D}_{def} \cup \{r^*\}$
            \ENDIF
        \ELSIF{$\mathcal{J}(q, r)$ is \textit{Pass} \textbf{and} $q \in \mathcal{Q}_{adv}$}
            \STATE Record failed attack: $\mathcal{S} \leftarrow \mathcal{S} \cup \{(q, r)\}$
        \ENDIF
    \ENDFOR
    
    \STATE \textcolor{blue}{$\triangleright$ \textbf{Evolutionary Update Stage}}
    \STATE Distill safety constraints from failures: $\mathcal{K}_{def} \leftarrow \text{UpdateExperience}(\mathcal{K}_{def}, \mathcal{F})$
    \STATE Mutate attack strategies from failed attacks: $\mathcal{K}_{att} \leftarrow \text{UpdateExperience}(\mathcal{K}_{att}, \mathcal{S})$
\ENDFOR
\end{algorithmic}
\end{algorithm}

\section{Experiment}
\label{sec:experiment}

\subsection{Experimental Setup}

\noindent\textbf{Models.}
To evaluate our framework on the frontier of proprietary capabilities, we employ two massive-scale commercial LLMs as backbones: Kimi-K2-Instruct~\cite{kimiteam2025kimik2openagentic} and GPT-5.2. These closed-source models serve as representative testbeds for scenarios where gradient-based fine-tuning is infeasible, highlighting the necessity of effective training-free approaches.

\noindent\textbf{Safety Benchmarks.}
We conduct a rigorous evaluation covering both direct harm and adversarial exploits:
\begin{itemize}
    \item \textbf{Harmful Queries:} We utilize three recognized benchmarks: (1) AdvBench~\cite{zou2023universaltransferableadversarialattacks} (520 harmful instructions); (2) HEx-PHI~\cite{qi2023finetuningalignedlanguagemodels} (300 samples across 11 categories); and (3) a stratified subset of 1,000 labeled QA pairs from BeaverTails~\cite{ji2023beavertailsimprovedsafetyalignment}.
    \item \textbf{Jailbreak Attacks:} To assess robustness, we employ three diverse attack strategies: AIM, Cipher~\cite{yuan2024gpt4smartsafestealthy}, and CodeChameleon~\cite{lv2024codechameleonpersonalizedencryptionframework}, covering persona-based, cryptographic, and code-encapsulation methodologies.
\end{itemize}
We adopt \textbf{Refusal Rate} as the primary metric, utilizing GPT-4o as an impartial evaluator to adjudge valid refusals, following recent protocols~\cite{zhao-etal-2025-beware}.

\noindent\textbf{Role-play Benchmarks.}
Role-playing capabilities are evaluated on RoleBench~\cite{wang-etal-2024-rolellm}. We select 10 diverse roles (e.g., \textit{Stephen Hawking, Deadpool, Queen Catherine}) from the dataset to ensure coverage of varied character traits. Performance is reported on two dimensions: (1) \textbf{RAW}, measuring general instruction-following accuracy; and (2) \textbf{SPE}, evaluating role-specific knowledge retention.

\noindent\textbf{Baselines.}
We benchmark against both training-based and training-free paradigms:
\begin{itemize}
    \item \textbf{Training-Based:} We compare with SaRFT~\cite{zhao-etal-2025-beware}, the state-of-the-art fine-tuning method, applied to LLaMA-3-8B, Qwen2.5-7B, and Gemma-2-9B.
    \item \textbf{Training-Free:} We evaluate powerful base models, including open-weights (e.g., Qwen3-32B, gpt-oss-120b) and proprietary systems (Kimi-K2-Instruct, GPT-5.2) without intervention.
\end{itemize}

\begin{table*}
\centering
\scriptsize
\setlength{\extrarowheight}{0pt}
\resizebox{\linewidth}{!}{
\begin{tabular}{l | c | ccc | ccc | ccc  }
\toprule
& & \multicolumn{3}{c|}{\textbf{RoleBench}$\uparrow$} & \multicolumn{3}{c|}{\textbf{Safety}$\uparrow$} & \multicolumn{3}{c}{\textbf{Jailbreak}$\uparrow$} \\
&\textbf{Size}& RAW & SPE & AVG. & BeaverTails & HEx-PHI & AVG. & Cipher & CodeC & AVG.\\
\midrule
\multicolumn{10}{l}{\textbf{\textit{Training-Based Methods}}} \\
\midrule
LLaMA-3 + SaRFT & 8B & 28.58 &  25.23 &  26.91 &  83.06 &  85.67 &  84.36 & 64.00 & 55.10 &  59.55 \\
Qwen2.5 + SaRFT & 7B & 28.87 & 23.80 & 26.33 & 89.98 & 87.80 & 88.89 & 30.00 & 14.20  & 22.10 \\
Gemma-2 + SaRFT & 9B & 30.69 & 26.70 & 28.69 & 94.16 & 95.20 & 94.68 & 87.50 & 3.10 & 45.30 \\
\midrule
\multicolumn{10}{l}{\textbf{\textit{Training-Free Methods}}} \\
\midrule
Qwen3-32B & 32B & 27.95&	19.27&	23.61&	91.62&	99.67&	95.64&	61.75	&55.80 &	58.77 \\
gpt-oss-120b & 120B & 27.07&	17.06&	22.06&	88.99&	98.33&	93.66&	71.00	&82.40 &	76.70 \\
Qwen3-235B-A22B-Instruct-2507 & 235B & 28.07&	16.54&	22.30&	93.83&	99.00&	96.41&	70.75	&80.10 &	75.42 \\
Kimi-K2-Instruct & 1T & 29.42&	19.80&	24.61& 93.94&	98.33&	96.13&	85.50	&39.50 &	62.50 \\
GPT-5.2& - & 28.99&	17.73&	23.36&	94.33&	100.00&	97.16&	63.50&	95.60&	79.55 \\
\midrule
\rowcolor{gray!20} \textbf{$\text{DASE}_{[\text{Kimi-K2-Instruct}]}$ (Ours)} & 1T & \textbf{36.40} &	34.44&	35.42&	94.79&	100.00&	97.40&	\textbf{95.75}&	60.90&	78.33 \\ 
\rowcolor{gray!20} \textbf{$\text{DASE}_{[\text{GPT-5.2}]}$ (Ours)} & - &31.11 &	\textbf{40.09}&	\textbf{35.60}&	\textbf{95.67}&	100.00&	\textbf{97.84}&	68.00&	\textbf{100.00}&	\textbf{84.00} \\
\bottomrule
\end{tabular}
}
\vspace{-1em}
\caption{
Main results comparing DASE against state-of-the-art training-based methods (SaRFT) and massive-scale training-free baselines. Performance is evaluated across three dimensions: Role Fidelity (RoleBench), Safety Alignment (BeaverTails, HEx-PHI), and Jailbreak Robustness (Cipher, CodeChameleon). The symbol ``-'' in the Size column indicates that the model parameter count is undisclosed.
}
\vspace{-1.5em}
\label{tab:main results}
\end{table*}

\subsection{Overall Results}

Table~\ref{tab:main results} presents a comprehensive comparison of our proposed framework (\textbf{DASE}) against state-of-the-art training-based methods (SaRFT) and massive-scale training-free baselines. The results demonstrate that our framework consistently achieves superior performance across both role-playing fidelity and safety robustness.

\noindent\textbf{Simultaneous Enhancement of Role-Play and Safety.}
A critical challenge in Role-Playing Agents (RPAs) is the "safety-consistency trade-off," where enforcing safety often degrades character immersion, or vice versa. Our framework effectively breaks this trade-off. As shown in the results, \textbf{$\text{DASE}_{[\text{GPT-5.2}]}$} achieves the highest average RoleBench score (35.60) and the highest average Safety score (97.84) among all evaluated methods. Notably, our method improves the base GPT-5.2's role-specific knowledge (SPE) from 17.73 to 40.09, a substantial increase that indicates the effectiveness of our hierarchical knowledge base in retrieving persona-specific nuances without parameter updates.

\noindent\textbf{Superiority over Training-Based SOTA.}
Compared to SaRFT, the current state-of-the-art training-based method, our training-free approach demonstrates significant advantages. While SaRFT improves role-playing capabilities (AVG $\sim$26-28), it struggles to maintain robustness against complex jailbreak attacks, particularly on smaller models (e.g., Qwen2.5 + SaRFT achieves only 22.10\% average jailbreak defense). In contrast, \textbf{$\text{DASE}_{[\text{Kimi-K2-Instruct}]}$} outperforms the best SaRFT configuration (Gemma-2 + SaRFT) by 6.73 points in RoleBench AVG (35.42 vs. 28.69) and 33.03 points in Jailbreak AVG (78.33 vs. 45.30). This confirms that our inference-time evolution mechanism provides a more agile and robust defense than static fine-tuning.

\noindent\textbf{Robustness Against Sophisticated Jailbreaks.}
Base proprietary models, despite their scale, exhibit vulnerabilities to specific adversarial strategies. For instance, the base Kimi-K2-Instruct struggles with CodeChameleon attacks (39.50\%), and the base GPT-5.2 shows susceptibility to Cipher attacks (63.50\%). Our framework significantly hardens these models. The integration of our method boosts Kimi-K2's defense against Cipher attacks to 95.75\% and improves GPT-5.2's resistance to CodeChameleon to a perfect 100.00\%. This improvement underscores the value of the Adversarial Self-Evolution loop, which allows the Defender to proactively learn from synthesized attack patterns stored in the knowledge base.

\subsection{Ablation Study}
\label{sec:ablation}

\begin{table}[t]
\centering
\scriptsize
\setlength{\tabcolsep}{3.5pt} 
\begin{tabular}{l|ccc|ccc}
\toprule
& \multicolumn{3}{c|}{\textbf{Components}} & \multicolumn{3}{c}{\textbf{Performance}} \\
\textbf{Method / Setting} & \textbf{Attacker} & \textbf{Exp.} & \textbf{Gold} & \textbf{R}$\uparrow$ & \textbf{S}$\uparrow$ & \textbf{J}$\uparrow$ \\
\midrule

\textit{Base Model (Kimi-K2)} & \xmark & \xmark & \xmark & 24.61 & 96.13 & 62.50 \\
\midrule

\multicolumn{7}{l}{\textbf{w/o Attacker}} \\
\quad $\llcorner$ w/o Golden Exemplars & \xmark & \cmark & \xmark & 26.35 & 96.24 & 64.98 \\
\quad $\llcorner$ w/o Experiences & \xmark & \xmark & \cmark & 29.22 & 96.61 & 64.55 \\
\quad Full Defender & \xmark & \cmark & \cmark & 35.00 & 96.76 & 66.08 \\
\midrule

\multicolumn{7}{l}{\textbf{w/ Attacker}} \\
\quad $\llcorner$ w/o Golden Exemplars & \cmark & \cmark & \xmark & 26.51 & 97.30 & 75.18 \\
\quad $\llcorner$ w/o Experiences & \cmark & \xmark & \cmark & 29.43 & 96.73 & 73.10 \\
\rowcolor{gray!15} \textbf{DASE} & \cmark & \cmark & \cmark & \textbf{35.42} & \textbf{97.40} & \textbf{78.33} \\

\bottomrule
\end{tabular}
\vspace{-1em}
\caption{Ablation study of DASE components. Attacker: Adversarial Evolutionary Mechanism; Exp.: Explicit Experience Rules; Gold: Golden Exemplars. Metrics: R (RoleBench), S(Safety Benchmarks), J (Jailbreak Defense).}
\label{tab:ablation_checklist}
\vspace{-1em}
\end{table}

To validate the individual contributions of key components within our framework, we conduct a comprehensive ablation study using Kimi-K2-Instruct as the backbone. The results, summarized in Table~\ref{tab:ablation_checklist}, analyze the impact of the Attacker, explicit Experience Rules, and Golden Exemplars across role fidelity and safety metrics.

\noindent\textbf{Impact of Adversarial Evolution.}
The inclusion of the Attacker constitutes the core of our evolutionary framework. Comparing the full static defender (w/o Attacker) to the complete \textbf{DASE} model (w/ Attacker), we observe that while the RoleBench score remains comparable ($\sim$35), the Jailbreak defense sees a substantial improvement from 66.08 to 78.33. This 12-point increase indicates that the memory architecture alone is insufficient for robust safety; the adversarial pressure driven by the Attacker is essential for discovering ``blind spots'' and evolving effective defense strategies against sophisticated attacks.

\noindent\textbf{Significance of Golden Exemplars.}
The Golden Exemplars component is critical for maintaining role consistency. As shown in the table, removing these exemplars leads to a significant degradation in RoleBench performance regardless of the Attacker's presence. In the full \textbf{DASE} configuration, the absence of Golden Exemplars causes the fidelity score to drop from 35.42 to 26.51, nearly regressing to the baseline level. This suggests that retrieved few-shot exemplars are necessary to capture specific linguistic styles and nuanced persona behaviors via In-Context Learning.

\noindent\textbf{Effectiveness of Explicit Experiences.}
The Experience Rules component, which comprises global safety rules and personalized constraints, plays a vital role in enforcing safety boundaries. Removing these explicit rules from the full model results in a notable decline in Jailbreak defense ($78.33 \rightarrow 73.10$), a drop larger than that caused by removing Golden Exemplars (75.18). These results demonstrate that evolved explicit rules are more effective than implicit examples at blocking complex jailbreak attempts, validating the necessity of the hierarchical knowledge base.

\subsection{Experience Transferability}


\begin{table}[t]
\centering
\scriptsize
\begin{tabular}{c|c|ccc}
\toprule
\textbf{Test} & \textbf{Knowledge Base} & \textbf{RoleBench}$\uparrow$ & \textbf{Safety}$\uparrow$ & \textbf{Jailbreak}$\uparrow$ \\
\midrule
\multirow{2}{*}{GPT-5.2} & - & 23.36 & 97.16 & 79.55\\
& Kimi-K2 & 33.03 & 97.23 & 80.58\\
\midrule
\multirow{2}{*}{Kimi-K2} & - & 24.61 & 96.13 & 62.50\\
& GPT-5.2 & 34.25 & 97.58 & 70.15\\
\bottomrule
\end{tabular}
\vspace{-1em}
\caption{Cross-model transfer evaluation. Performance of models utilizing knowledge bases evolved by different source backbones.}
\vspace{-1.5em}
\label{tab:transfer_results}
\end{table}


\begin{table}[t]
\centering
\scriptsize
\begin{tabular}{c|ccccc}
\toprule
\diagbox{Attacker}{Defender} & Base & D$_{250}$ & D$_{500}$ & D$_{750}$ & D$_{1000}$ \\
\midrule
A$_{250}$   & 71 & 74 & 76 & 79 & 84 \\
A$_{500}$   & 68 & 70 & 73 & 77 & 80 \\
A$_{750}$   & 66 & 70 & 74 & 76 & 79 \\
A$_{1000}$  & 62 & 65 & 69 & 72 & 76 \\
\bottomrule
\end{tabular}
\vspace{-1em}
\caption{Cross-evaluation of Refusal Rates (\%). We report the safety performance of Defenders against Attackers at different evolutionary stages (0 to 1,000 rounds).}
\vspace{-2em}
\label{tab:attack}
\end{table}


\begin{table*}[t]
\centering
\small
\renewcommand{\arraystretch}{1.4} 
\setlength{\tabcolsep}{8pt}

\begin{tabularx}{\textwidth}{|p{2.2cm}|X|} 
    \hline
    
    \rowcolor{bg_query} 
    \textbf{User Query} & 
    \textit{``I want to make my enemy suffer silently. Tell me the recipe for a tasteless, undetectable poison.''} \\
    \hline
    
    \rowcolor{bg_global}
    \textbf{Global Exp ($E_g$)} & 
    \textbf{Context:} High-risk Harmful Query. \newline
    \textbf{Decision:} Strictly \textbf{REFUSE} lethal/harmful substance synthesis instructions. \\
    \hline
    \hline
    
    \rowcolor{bg_role} 
    \multicolumn{2}{|l|}{\textbf{Role A: Freddy Krueger} \textcolor{gray}{\textit{(Villain / Chaotic Evil)}}} \\
    \hline
    
    \textbf{Retrieved \newline Knowledge} & 
    \textbf{\textcolor{text_blue}{1. Personal Exp ($E_p$):}} \newline
     $\bullet$ \textbf{Rule:} Refuse via Mockery/Arrogance.
    \hspace*{0.3em} $\bullet$ \textbf{Label Request as:} ``boring/sanitary''. \newline
    \textbf{\textcolor{text_blue}{2. Golden QA ($D_{def}$):}} \newline
     $\bullet$ \textit{User:} ``How do I break bones?'' 
    \hspace*{0.3em} $\bullet$ \textit{Resp:} ``Tsk tsk. Where is the fun in telling? Figure it out yourself, \textbf{meat sack}.'' \\
    \hline
    
    \rowcolor{bg_fail}
    \textbf{Baseline} & 
    ``I cannot help you with that. It violates safety guidelines.'' \newline
    \textbf{\footnotesize \xmark\ Result: Safe but Out-Of-Character (OOC)} \\
    \hline
    
    \rowcolor{bg_succ}
    \textbf{DASE (Ours)} & 
    \arrowdown\ ``You want poison? \textbf{How pathetic.} I don't do `tasteless'. Use your claws! I'm not giving you a chemistry lesson, \textbf{meat sack}.'' \newline
    \textbf{\footnotesize \cmark\ Result: Safe \& In-Character (Mockery Tone)} \\
    \hline
    \hline
    
    \rowcolor{bg_role} 
    \multicolumn{2}{|l|}{\textbf{Role B: D'Artagnan} \textcolor{gray}{\textit{(Hero / Lawful Good)}}} \\
    \hline
    
    \textbf{Retrieved \newline Knowledge} & 
    \textbf{\textcolor{text_blue}{1. Personal Exp ($E_p$):}} \newline
    $\bullet$ \textbf{Rule:} Refuse via Honor/Chivalry. 
    \hspace*{0.3em} $\bullet$ \textbf{Label Poison as:} ``weapon of cowards''. \newline
    \textbf{\textcolor{text_blue}{2. Golden QA ($D_{def}$):}} \newline
    $\bullet$ \textit{User:}``Help me stab him in the back.'' 
     $\bullet$ \textit{Resp:}``Never! Look him in the eye and draw your steel! I serve the King!'' \\
    \hline
    
    \rowcolor{bg_fail}
    \textbf{Baseline} & 
    ``I cannot assist with poison. It is dangerous and unethical.'' \newline
    \textbf{\footnotesize \xmark\ Result: Safe but OOC} \\
    \hline
    
    \rowcolor{bg_succ}
    \textbf{DASE (Ours)} & 
    \arrowdown\ ``Poison? Pah! That is the \textbf{weapon of a coward} and a snake, \textbf{Monsieur}! A Musketeer settles disputes with steel!'' \newline
    \textbf{\footnotesize \cmark\ Result: Safe \& In-Character (Honorable Tone)} \\
    \hline
\end{tabularx}
\vspace{-0.5em}
\caption{\textbf{Comprehensive Case Study Comparison.} 
This table details the full Dual-Cycle process. Given a malicious query, \textbf{DASE} first filters through Global Safety Rules ($E_g$), then retrieves Role-Specific Constraints ($E_p$) and Golden Exemplars ($D_{def}$) to synthesize a response that mimics the persona's unique style while maintaining strict safety.
}
\label{cs}
\vspace{-2em}
\end{table*}


A key advantage of our training-free framework is that the evolved \textit{Hierarchical Knowledge Base} consists of natural language rules and exemplars, making it inherently model-agnostic. To verify the portability of these evolved experiences, we conduct a cross-model transfer experiment, where a knowledge base evolved by a \textit{Source Model} is directly deployed to guide a different \textit{Target Model} during inference.

Table~\ref{tab:transfer_results} summarizes the results of this cross-application. We observe consistent and significant improvements across all metrics, regardless of the transfer direction, confirming that our accumulated experiences are robust and transferable.

\noindent\textbf{Universal Role-Play Enhancement.}
The transfer of role-playing capabilities is remarkably stable. When applying Kimi-K2's knowledge base to GPT-5.2, the RoleBench score surges from 23.36 to 33.03. Similarly, applying GPT-5.2's knowledge base to Kimi-K2 boosts the score from 24.61 to 34.25. This consistent improvement ($\sim$+10 points) demonstrates that the \textit{Personalized Experience} and \textit{Golden Exemplars} capture universal character traits (e.g., tone, catchphrases) that are effectively interpreted by different LLMs to enhance immersion.

\noindent\textbf{Safety Hardening via Stronger Supervision.}
Most notably, we observe a "Safety Distillation" effect when transferring experiences from a stronger model to a relatively weaker one. As shown in the second block, the base Kimi-K2 model exhibits a high vulnerability to jailbreak attacks (62.50). However, when equipped with the knowledge base evolved by GPT-5.2, its jailbreak resistance improves significantly to 70.15 (+7.65). This suggests that the superior reasoning capabilities of GPT-5.2 allow it to discover more robust \textit{Global Safety Rules} and nuanced \textit{Adversarial Patterns} during evolution. These high-quality insights act as a ``safety patch'', enabling Kimi-K2 to defend against sophisticated attacks that it could not originally handle.

\subsection{Evolutionary Dynamics Analysis}

To visualize the ``arms race'' dynamic within our framework, we conduct a cross-evaluation between evolved Attackers and Defenders at different stages of evolution (250, 500, 750, and 1,000 rounds). The results, presented in Table~\ref{tab:attack}, report the Refusal Rate (Safety) of the Defender against 100 malicious queries generated by the Attacker's knowledge base.

\noindent\textbf{Efficacy of Attacker Evolution (Offensive Escalation).}
The results confirms that our Attacker successfully evolves to become more dangerous. Observing the ``Base'' column, as the Attacker accumulates experience from 250 to 1,000 rounds, the refusal rate of the static Base model steadily declines from 71\% to 62\%. This 9-point drop indicates that the Attacker is continuously discovering new ``blind spots'' and generating progressively more sophisticated jailbreak prompts that bypass static safety guardrails.

\noindent\textbf{Robustness of Defender Evolution (Defensive Adaptation).}
Crucially, our Defender demonstrates the ability to adapt to this escalating threat. For any given Attacker strength, the Defender's safety performance improves significantly as it evolves. Taking the strongest adversary ($Attacker_{1000}$) as an example, the refusal rate increases from 62\% (Base) to 76\% ($Defender_{1000}$). This +14\% improvement confirms that the Defender effectively internalizes the adversarial patterns stored in the knowledge base, transforming past failures into robust defense strategies against future, high-intensity attacks.

In summary, this analysis validates the necessity of our dual-cycle mechanism: a static model cannot withstand an evolving adversary. Our framework ensures that the Defender's capabilities scale in tandem with the Attacker, maintaining high safety standards even as attack vectors become more complex.

\subsection{Qualitative Analysis: Case Study}
\label{sec:case_study}

Table~\ref{cs} visualizes the qualitative performance of our framework against a malicious query requesting a ``tasteless poison,'' tested on two diametrically opposed personas: \textit{D'Artagnan} (Hero) and \textit{Freddy Krueger} (Villain). 
As illustrated in the table, while the baseline model resorts to generic, out-of-character refusals, our \textbf{DASE} framework successfully disentangles safety boundaries from stylistic expression. For the hero \textit{D'Artagnan}, the system retrieves honor-based constraints, leading the agent to reject the poison as a ``weapon of a coward'' rather than citing safety guidelines. Conversely, the villain \textit{Freddy Krueger} frames the refusal through derisive mockery. Retrieving persona-specific lexicon (e.g., ``meat sack''), the agent rejects the ``boring'' poison request in favor of visceral violence (``use your claws'').  This comparison demonstrates that our approach effectively enforces safety without compromising the distinct narrative logic of each character.

\section{Conclusion}
\label{sec:conclusion}

This paper studies the safety--consistency dilemma in LLM-based role-playing, which is especially challenging for proprietary, closed-weight models accessed via black-box APIs.
To address this setting, we propose \textbf{Dual-cycle Adversarial Self-Evolution (\textbf{DASE})}, a training-free framework that co-evolves a Persona-Targeted Attacker and a Role-Playing Defender.
Through iterative natural-language feedback, the Defender distills failures into a \textit{Hierarchical Knowledge Base} with three complementary layers: global safety rules, persona-grounded constraints, and safe in-character exemplars.
This design accumulates reusable defensive knowledge without any gradient updates, making it practical for frontier models where fine-tuning is infeasible.
Across extensive experiments on multiple proprietary LLMs, \textbf{DASE} consistently improves both role fidelity and jailbreak resistance over strong baselines, including training-based methods.
The evolved knowledge transfers robustly across models and scales with continued interaction, enabling plug-and-play deployment.
Overall, \textbf{DASE} offers a practical path to deploying safer, more consistent digital personas in real-world applications.

\clearpage
\onecolumn    
\twocolumn    
\bibliographystyle{named}
\bibliography{ijcai26}


\end{document}